\documentclass[preprint,authoryear,12pt]{elsarticle}

\usepackage{amssymb}

\journal{Artificial Intelligence}

\begin{document}

\begin{frontmatter}



\title{Are Minds Computable?}


\author{Carlos Gershenson}

\address{Departamento de Ciencias de la Computaci\'on\\
Instituto de Investigaciones en Matem\'aticas Aplicadas y en Sistemas \\
Universidad Nacional Aut\'onoma de M\'exico\\
A.P. 20-726, 01000 M\'exico D.F. M\'exico
\ead{cgg@unam.mx}
\ead[url]{http://turing.iimas.unam.mx/~cgg}
}

\begin{abstract}
This essay explores the limits of Turing machines concerning the modeling of minds and suggests alternatives to go beyond those limits.
\end{abstract}


\end{frontmatter}



One of the main open questions in science and philosophy since ancient times has been how bodies relate to minds. Ren\'e Descartes proposed his famous dualism to ontologically separate the physical (\emph{res extensa}) from the mental (\emph{res cogitans}). However, with Cartesian dualism one cannot relate bodies and minds. This has led many people to propose ways of reducing minds to the physical realm \citep[e.g.][]{Ashby1947}. 

Alan Turing \citeyearpar{Turing:1950} was also interested in describing minds using only mechanical principles. Universal Turing machines \citep{Turing:1937} were proposed to decide whether a \emph{mechanism} could produced a desired computation. Thus, one can transform the mind-body relation into the mind-mechanism relation \citep{Johnson:1987,Boden:1988}, avoiding the ontological part of the mind-body problem. This can be formulated with the question: \emph{Can universal Turing machines compute a process similar to a mind?} Turing himself proposed his famous test to decide this \citep{Turing:1950,Proudfoot:2011}. This question is relevant for Artificial Intelligence, since a large part of A.I. research models aspects of human minds on machines with an architecture inspired by Turing machines.

Stuart Kauffman \citeyearpar{Kauffman:2012} has recently argued that Turing machines are not able to compute minds. After noting several limitations of Turing machines, Kauffman---following others such as Penrose \citeyearpar{Penrose:1989}---suggests an alternative in quantum mechanics. Here I defend that, even when Turing machines are not sufficient for computing minds,  quantum mechanics is not required to provide a more viable explanation. The crucial difference between Turing machines and human minds lies in \emph{interactions} \citep{Gershenson:2011e}.

Universal Turing machines (UTMs) have been a central concept in computer science and A.I. \citep{Denning:2010}. UTMs are used to define computable functions. They are one of the theoretical foundations of modern computers. Still, they have limitations. Since A.I. has been working with computers based on the concept of a UTM, it is affected by the same limitations. Alternatives to classical A.I. \citep[e.g.][]{RumelhartEtAl1986,Brooks:1991} also suffer from these limitations, since these do not depend on representations \citep{Gershenson2004}.

The main limitations of UTMs---related to modeling minds---are two: 
\begin{enumerate}
\item UTMs are closed.
\item UTMs compute---i.e. produce an output---only once they halt. 
\end{enumerate}

UTMs are closed because once an initial condition and a program are set, the result will be deterministic. However, there are many phenomena where data (in the tape or program) changes at runtime. One example of this can be seen with coupled Turing machines \citep{Copeland:1997,Copeland:1999}. Certainly, if the interactions and updating scheme between Turing machines (where the computation of one machine affect the data of another machine) are deterministic and defined precisely, these could be modeled by a UTM. Nevertheless, in most cases these are only known \emph{a posteriori}, because of computational irreducibility \citep{Wolfram:2002}. Thus, one cannot define the output of an ``open" Turing machine, since the inputs during runtime are unknown \emph{a priori}. As opposed to UTMs, minds are constantly receiving inputs.

Computations carried out by UTMs have to halt before an output can be extracted from them. However, there are many computations that do not halt \citep{Wegner:1998,Denning:2011}, for example, biological computation \citep{Mitchell:2011} is not about computing a function, but a continuous information processing that sustains living systems \citep{MaturanaVarela1980}. Minds also fall within this category. A mind does not compute a function and halts. A mind is constantly processing information.

From the above arguments, it can be concluded that UTMs cannot compute minds. Thus, A.I. as it stands now cannot model minds completely. Still, this does not imply that minds cannot be computed. Now the question is: \emph{Are there (non-Turing) mechanisms capable of computing a process similar to a mind?} If we see computation as transformation of information \citep{Gershenson:2007}, then human minds are already computing processes. Thus, the answer is affirmative---human minds can be seen as computers---but not very useful. Minds are computable, but how?. A more pragmatic framing of the question would be: \emph{How can we describe computations of processes similar to minds?}. This description should enable us to understand better cognitive processes and also to build more sophisticated artificial cognitive systems.

As it was argued above, \emph{interactions} \citep{Gershenson:2011e} are the missing element in UTMs. Probably this is due to a reductionist bias in science. Classical science, since the times of Galileo, Descartes, Newton and Laplace, has tried to simplify and separate in order to predict \citep{Kauffman2000}. This is understandable, since traditional modeling is limited by the amount of information included in the model. This has naturally led to neglecting interactions. However, modern computers---a product of reductionist science---have enabled us to include much more information in our models, opening the possibility to include interactions. 

Computing with interactions \citep{Wegner:1998}  implies models of computation that are open and that do not halt. Interestingly, modern computers---while based on UTMs---are able to perform interactive computation. It should be noted that Turing-computability---which is theoretical---is different from practical computability (in a physical computer). For example, there are Turing-computable functions that are not computable in practice simply because there is not enough time and/or memory in the universe. On the other hand, computers can compute  non-Turing-computable functions, such as halting functions with the aid of an ``oracle" (a posteriori). Interactions are another example, since computers may receive inputs while processing information and may not halt. This is positive for A.I. since it implies that modeling minds with current computers does not imply a hardware challenge. 

The remaining challenge is related to reductionism. By necessity, many A.I. systems rely on interactions, be it with humans or with other artificial systems. Examples include conversation systems \citep{Pineda:2007} and agents for the Semantic Web \citep{Hendler:2001}. However, even these systems simplify modeling to reductionist processes, where allowed interactions are defined beforehand, and computations are well defined. This contrasts with natural systems---humans included---where there is a constant interaction with the environment \citep{Clark1997} that is not predefined. The ability to respond to unforeseen circumstances seems an obstacle for A.I. Still, there are several examples of adaptive systems that go beyond this limitation \citep[e.g.][]{Braitenberg:1986,Bongard:2006}. 

A.I. can describe minds in terms of computation only if this is dynamic (changing in runtime) and continuous (non halting). This will enable artificial systems to respond to interactions which can occur at any moment and do not stop occurring. 

To achieve dynamic computation, systems should be able to be interrupted and modified. New states should be able to be created and old states should be able to be deleted.

To achieve continuous computation, systems should be programmed  with the goal of constant interaction, as opposed to producing a set of outputs. Systems should not deliver results. Systems should deliver processes.

Dynamic, continuous computing will certainly be less predictable than traditional computing. Still, it will have the possibility of being more creative and adaptive, since new solutions will be constantly sought as novel situations arise.

\bibliographystyle{elsarticle-num-names}
\bibliography{carlos,sos,computing,COG,complex,RBN}

\end{document}